# Improving healthcare access management by predicting patient no-show behaviour[1]

David Barrera Ferro *

barrera-o@javeriana.edu.co

Southampton Business School, University of Southampton. Southampton, United Kingdom.

Industrial Engineering Department, Pontificia Universidad Javeriana. Bogotá, Colombia.

Sally Brailsford

S.C.Brailsford@soton.ac.uk

Southampton Business School, University of Southampton. Southampton, United Kingdom.

Cristian Bravo

cbravoro@uwo.ca

Department of Statistical and Actuarial Sciences, The University of Western Ontario. London, ON, Canada.

Honora Smith

Honora.Smith@soton.ac.uk

Mathematical Sciences, University of Southampton. Southampton, United Kingdom.

* Corresponding Author

---

**Abstract**

Low attendance levels in medical appointments have been associated with poor health outcomes and efficiency problems for service providers. To address this problem, healthcare managers could aim at improving attendance levels or minimizing the operational impact of no-shows by adapting resource allocation policies. However, given the uncertainty of patient behaviour, generating relevant information regarding no-show probabilities could support the decision-making process for both approaches. In this context many researchers have used multiple regression models to identify patient and appointment characteristics than can be used as good predictors for no-show probabilities. This work develops a Decision Support System (DSS) to support the implementation of strategies to encourage attendance, for a preventive care program targeted at underserved communities in Bogotá, Colombia. Our contribution to literature is threefold. Firstly, we assess the effectiveness of different machine learning approaches to improve the accuracy of regression models. In particular, Random Forest and Neural Networks are used to model the problem accounting for non-linearity and variable interactions. Secondly, we propose a novel use of Layer-wise Relevance Propagation in order to improve the explainability of neural network predictions and obtain insights from the modelling step. Thirdly, we identify variables explaining no-show probabilities in a developing context and study its policy implications and potential for improving healthcare access. In addition to quantifying relationships reported in previous studies, we find that income and neighbourhood crime statistics affect no-show probabilities. Our results will support patient prioritization in a pilot behavioural intervention and will inform appointment planning decisions.

**1. Introduction**

High no-show rates are a major issue for health systems. On the one hand, there is a link between low attendance levels and poor health outcomes: consequences include delays in diagnosis and initiation of treatment [1], increased premature mortality rates [2] and increased use of emergency services [3], among others. On the other hand, high no-show rates reduce efficiency for service providers. When a patient fails to keep an appointment, it usually results in a vacant slot that might have been used by another patient [4], increases cost of care [5] and generates idle time for both physicians and consultancy rooms [6]. Consequently, there is a growing interest from the healthcare community in understanding the determinants of no-show behaviour [7] and minimizing its impact [8].

Two main approaches can be used to deal with no-shows in healthcare settings: to improve attendance levels or to minimize impact. The first approach is premised on the idea that it is possible to change patient behaviour. Strategies such as phone reminders [9] and education programs [5] have been successfully implemented in different contexts. According to Zebina et al. [1] a key element in this approach is to be able to correctly identify the patients that should be targeted with each strategy. In contrast, the underlying assumption of the second approach is that such changes in behaviour are unlikely to be achieved, and thus the objective is to minimize the impact of no-shows, e.g. by improving the decision-making process regarding resource allocation and scheduling [8]. Given the uncertainty of patient behaviour, generating relevant information concerning no-show probabilities could improve the results of both approaches [10,11].

Good estimates of attendance levels have the potential to improve policy evaluation [12], to minimize undesired effects of resource allocation practices (such as overbooking) [13] and to inform identification of influential stakeholders. This is particularly important considering that one of the main objectives is to generate information that can be used to improve management practices [14]. However, there has been very little discussion in the literature on the trade-offs between using 'black box' approaches (sophisticated analytical methods that would be incomprehensible to most healthcare managers) and more easily interpretable but less accurate approaches such as regression models [15].

The context of the study described in this paper is a primary healthcare program targeted at underserved communities in Bogotá, Colombia. Under this program, community workers make home visits to assess health-risk levels and then, if required, schedule medical appointments. Over the two years 2017-2018 the program coverage grew considerably but unfortunately no-show rates for the scheduled medical appointments also grew, reaching levels of 35% and above. Consequently, there was pressure to improve planning processes and use resources more efficiently. In this work, we develop a Decision Support System (DSS) to support the implementation of strategies to reduce no-show rates in this program. It has been argued that in complex problems with multiple variables and fragmented data, information systems have potential to improve resource allocation practices [16] and increase system performance [17]. Hence, the DSS will use routinely collected data and Machine Learning (ML) methods to classify patients in three risk categories (low, medium and high) in terms of their individual no-show probability. Any intervention to encourage attendance will incur a cost, and hence it is more cost-effective to target such interventions at medium and high risk patients.

We adopt the Design Science Research (DSR) approach of Peffers et al. [25] to describe the development of the DSS and address three research questions:

- How reliably can routinely collected data on patient and appointment characteristics be used to predict no-show probabilities?
- What is the added value, in terms of AUROC (Area Under the Receiver Operating Characteristic curve), of using different ML approaches to predict no-show probabilities?
- How might insights obtained from these classification models be used in practice to reduce no-shows?

The paper is structured as follows. Section 2 presents recent work in no-show prediction, both in terms of the variables used and the modelling approach. Section 3 presents our DSR approach, describing the problem definition, the proposed solution, the design, and the demonstration and evaluation phases. Section 4 presents a descriptive analysis of the available data. Section 5 discusses the results and how these no-show risk classifications could be used in practice. Finally, Section 6 presents some general reflections.

**2. Related work**

Dantas et al. [7] reviewed studies about no-show prediction in health care, published between 1980 and 2016, and classified each reference considering the prediction variables, the modelling approach and the context of the application. The authors found that, over the last ten years, most studies use Multiple Logistic Regression models to quantify relationship between patient characteristics and no-show behaviour. Additionally, despite the highest no-show rates in the world being in Africa (43%), South America (28%) and Asia (25%), this problem was found to have been mainly studied in developed countries.

No-show behaviour in primary care appointments has been widely studied [7]. At least two features can been argued to explain the scientific interest: firstly, since primary care services are designed to serve large populations, the economic impact of inefficiencies may be greater than in specialized low-coverage services [18]. Secondly, primary care patients are highly heterogeneous; thus, there is evidence that supports contradictory results regarding the impact that patient characteristics might have on no-show rates [19]. In this section, we discuss no-show studies that are related to primary care settings and were published after the review conducted by Dantas et al. [7]. Table 1 presents the modelling approach, sample size and main variables in each of the relevant studies.

To understand the impact of a particular feature on no-show rates, several studies have been conducted using large datasets. McComb et al. [20] find that the impact of lead time on no-show rates is greater among patients who cancel and were rescheduled. Ellis, Luther & Jenkins [21] conclude that reduced sleep consequence of the spring daylight savings change, increased no-show rates, suggesting seasonality on the patterns. Both studies present ways in which the results could be used to improve scheduling practices. More recently, Wallace et al. [3] conclude that patients with lower income and longer travel times to the medical facility are more likely to miss their appointments and Parker, Gaugler, Samus and Gliting [22] find lower no-show rates among care givers using adult day services.

A second approach is to identify patient- and practice-related factors that predict no-show probabilities, using regression models. Ellis et al. [19] find that age, socioeconomic status and lead times are good predictors for repeated non-attendance in Scotland. Analysing data from hospitals in south-west England, French et al. [23] conclude that children from higher deprivation areas are more likely to miss their appointments. Goffman et al. [24] report that no-show history, age and having multiple appointments scheduled on the same day are good predictors for no-show rates among veterans in the United States. Ding et al. [25] discuss the need for designing different risk models for each medical service and facility in order to improve accuracy. Lastly, Tsai, Lee, Chiang, Chen, and Chen [26] find that patient gender, age and no-show history are good predictors in Taiwan.

In Design Science Research (DSR), knowledge can be descriptive (about the phenomena) or prescriptive (about the human-built artefacts) [27]. This paper adds to the existing body of research on no-show behaviour not only by its use of ML methods on routine health data (the artefact) but also through its focus on developing countries (the phenomena). According to Dantas et al. [7], between 2005 and 2016, most of the research uses regression models in order to predict no-show probabilities. However, recent studies explore the use of other machine learning techniques. Mohammadi, Wu, Turkan, Toscos and Doebbeling [28] analyse data from electronic health records, over a 3-year period, from Community Health Centres in Indianapolis. The authors implement logistic regression, neural networks and a Bayes classifier to predict no-show probabilities on a dataset containing 73,811 appointments. Unusually, the regression models and Bayes classifier perform better than neural networks, with AUROC values of 0.81, 0.86 and 0.66, respectively. Similarly, Topuz et al. [15] assess the effectiveness of Bayesian Belief Networks and propose an elastic net variable selection methodology. The authors conclude that there is a potential for machine learning methods to support improvements in management practices by providing accurate prediction of no-show probabilities. These studies make no mention of interpretability of these black-box models, a topic we tackle in this paper.

Lastly, while the same modelling methodology can be generalized to different countries, and the same independent variables have been used across different contexts and service settings, there is considerable variation between service delivery processes and the demographic and epidemiological population profiles in different countries. Each setting must be independently modelled to generate relevant information. Furthermore, previous studies in developing countries report specific features unique to that setting. Challenges such as low use of technology to centralize patient information [29,30], income inequality and job instability [31,32], long travel distances and poverty [33], access barriers to specialized care [34] and even low quality expectations [31] lead to different interactions between patients and service providers. Consequently, a systematic effort to develop prediction models and customize strategies for developing countries is required, and we do so in this work.

Table 1 No-show studies in primary care settings published since 2017

| Reference | Country | Sample size | Method | Appointment variables | | | | | Patient variables | | | | |
|---|---|---|---|---|---|---|---|---|---|---|---|---|---|
| | | | | Lead time | Day | Distance | Weather | Season | Gender | No-show history | Age | Race | Marital Status |
| [19] | United Kingdom | 9,177,054 | LR | X | | | | | X | | | | |
| [23] | United Kingdom | 2,488 | LR | | | | | | X | | | | |
| [24] | United States | 18,000,000 | LR | | | | | | X | | | | X |
| [20] | United States | 46,710 | Chi-squared | X | | | | | | X | | | |
| [25] | United States | 2,231,000 | LR | X | | | | | | X | X | | |
| [21] | United Kingdom | 7,351,597 | Chi-squared | | | | | X | | | | | |
| [28] | United States | 73,811 | LR/BC/NN | X | X | | | | X | X | | X | X |
| [15] | United States | 105,344 | BBN | X | X | X | | | | X | X | | |
| [3] | United States | 51,580 | LR | | | X | | X | | | X | X | |
| [22] | United States | 509 | LR | | | | | | | | | X | X |
| [26] | Taiwan | 2,132,577 | LR | X | X | | X | | X | X | X | | |

LR: Logistic regression, BC: Bayes Classifier, NN: Neural Networks, BBN: Bayes Behaviour Network

## 3. A Design Science Research Approach

In this section, we present a DSR approach to allow program managers to select patients who will participate in different behavioural interventions, by designing an ML-based DSS. Table 2 presents an overview as five of the six steps in the DSR methodology proposed by Peffers et al. [35].

Table 2 Methodology for Design Science Research (Peffers et al. [25])

| Phase | Our Study |
|---|---|
| Problem definition and motivation | To reduce no-show rates, associated with poor patient outcomes and inefficient use of resources |
| Objectives for a solution | To allow program managers to select patients who will participate in different behavioural interventions aimed at increasing attendance levels. |
| Design and development | We compare four ML modelling approaches and address three questions:<br>1. How reliably can routinely collected data on patient and appointment characteristics be used to predict no-show probabilities?<br>2. Which ML approach performs best, using the AUROC metric?<br>3. How might insights obtained from these classification models be used in practice to reduce no-shows? |
| Demonstration | Performance assessment using the average AUROC score of a 10-by-10 Cross validation. |
| Evaluation | Impact on the coverage and risk of an intervention when a classification algorithm is used |

3.1 Problem Definition, Scope and Context.

Broadly speaking, the national healthcare system in Colombia (*Sistema General de Seguridad en Salud*, SGSS) can be understood as a managed competition model with two insurance schemes: one contributory, covering people who are in formal employment, and one subsidized, covering people unable to pay [36]. Despite guaranteeing universal coverage, the SGSS faces constant challenges to improve service quality, increase efficiency and eliminate access barriers [12]. Recent studies have shown that these challenges primarily affect patients of lower socioeconomic status [37–39]. Therefore, the District Secretary of Health (*Secretaría Distrital de Salud*, SDS) in Bogotá instituted a program to eliminate access barriers affecting low-income patients in the city. The program consists of a group of community workers visiting patients, who are sparsely geographically distributed, to assess risks, quantify needs and define care routes within the health system.

The service process of the program can be summarized in three phases. First, patients are identified using existing databases from other social programs. Second, community workers make home visits and classify patients as high, medium or low risk. Finally, in the third phase, a primary care pathway is defined for each patient, according to their level of health risk of needing a service. For each high or medium risk patient, a first medical appointment is scheduled in one or more services. Table 3 presents the objective of the first appointment, in each service, as defined by the National Health Authority [40]. At the end of this phase, the barrier is considered to be overcome, and the patient is expected to start treatment using the services of the relevant insurance scheme. The city is divided into four clusters providing health services and, for each cluster, a team is in charge of the operational decisions of the program.

Table 3  Services and objectives of the program

| Service | Objective |
|---|---|
| Oral health (OH) | To assess oral health status and promote self-care. |
| Grow and Development (G&D) | To assess and follow up growth and development status among children. |
| Young Adult Program (YAP) | To assess health status and development risks. |
| Senior Program (SP) | To assess health status and identity major changes. |
| CCU Program | To increase early diagnosis of cervical cancer. |
| Breast Cancer Screening | To increase early diagnosis of breast cancer. |
| Family Planning | To provide relevant information and counselling. |
| Antenatal Care | To ensure timely access and improve health outcomes. |
| Emergencies | To control health risks that might endanger quality of life. |
| Visual Care | To assess health status. |

The SDS has defined three performance indicators to assess the operation of the program.

(i) The percentage of *effective* visits, i.e. where the patient was physically present at the registered address at the time of the visit, and the community worker was able to assess them and make a risk classification.

(ii) The percentage of appointments given within a target lead time. The designated health centre may not have the capacity to treat the patient within the required time limit, and in such cases the earliest appointment is given.

(iii) The percentage of attendance at appointments (i.e., the percentage of no-shows). These patients might enter the health system later via emergency departments due to complications of the identified risks.

By the end of 2018, program managers faced challenges with indicators (ii) and (iii). Only 30% of appointments met lead time targets and no-show rates reached levels of 35% in some services.

3.2 The proposed solution

In this context, different interventions could be used to modify patient behaviour and improve program performance. Phone reminders [41,42], education [5] and engagement programs [43], among others, have shown positive impact in decreasing no-show rates in different service contexts. However, such interventions are most cost-effective when patients are classified according to their no-show risk [5,9]. Therefore, SDS has identified the need to divide patients into three groups. Group A will contain 30% of the patients, due to economic and operational constraints, no additional action will be implemented for them. For Group B, 40% of patients, lower-cost technology-based interventions such as SMS reminders will be evaluated. Group C will contain the remaining 30% of patients. For these, personalized interventions such as engagement or education programs will be designed to improve attendance levels.

It has been argued that by introducing decision support tools, organizations can encourage reasoned thinking, reduce bias and improve decision quality [44]. Therefore, in this paper, we design a DSS to allow program managers to select patients who will participate in different behavioural interventions. Additionally, when adopting a DSR approach, there is a need to explicitly formulate a set of statements describing the goal, and the means to achieve it. These prescriptive statements are called design principles and are a distinctive characteristic of design knowledge [45]. After conducting a series of meetings with program managers at SDS and reviewing research papers dealing with no-show behaviour for healthcare appointments, Table 4 presents the resulting components of our design principle using the schema proposed by Gregor et al. [45]. Within this strategy, the success of a set of interventions relies on the quality of the classification. Then, we define two performance indicators that describe the accuracy. The first is the coverage, defined as the percentage of no-show patients that end up classified in Group C, and the second is the risk, defined as the percentage of such patients that end up classified in Group A. A good classification will have high coverage and low risk. In contrast, a random classification would have both coverage and risk equal to 30%.

Table 4 Components of the design principle

| Design principle | Our DSS |
| --- | --- |
| Aim, Implementer and User | To allow program managers to select patients who will participate in different behavioural interventions aimed at increasing attendance levels. |
| Context | A primary care program, in a developing country, with high no-show rates. |
| Mechanism | Predict individual no-show probabilities using ML techniques. |
| Rationale | Behavioural interventions are most cost-effective when patients are classified according to their no-show risk. |

3.3 Design and development: the tested modelling approaches

The first step in the process is the quantification of linear relationships between variables and no-show probabilities. Ordinary least squares (OLS) estimation is widely used to that end. However, Tibshirani [46] analyses two major drawbacks of OLS: accuracy and interpretation. Since OLS estimates have large variance, setting some coefficients to zero contributes to overcoming both limitations. Therefore, Tibshirani proposes the LASSO regression model, which minimizes the residual sum of squares and ensures that the sum of the absolute value of the coefficients is less than some chosen value. Recent applications of this model include forecasting [47] and classification problems [48]. We have used Scikit-Learn's logistic regression CV implementation, setting the alpha value to 0 so as to not use ridge regression, and all other parameters have been left at their default values [49]. Additionally, we perform a parametric analysis on the penalty constant of the model. For each service, thirty values are tested (10 values for each of the following three intervals: (0-0.1], (0.1-1], (1-10]) selecting the constant under which the AUROC is stable (i.e. its improvement is marginal) and the prediction depends on the minimum number of variables. Then, for each input variable, we interpret the average coefficients of a 10-by-10 CV. These results are used to inform feature selection for both the RF and the NN.

Although logistic regression (LR) models are highly interpretable (i.e. understandable by non-experts), they may not be useful in contexts where the relationships between variables are nonlinear [50]. For those cases, tree-based ensemble algorithms have shown good performance and modelling flexibility [51]. Tree classifiers split the data set according to a criterion maximizing separation; the result is a tree-like structure [52]. An ensemble of tree predictors, where each tree depends on the values of a random sample of both cases and variables, is called a Random Forest [53]. For classification problems, RFs help to overcome the risk of overfitting, are less sensitive to outliers, and eliminate the need of pruning [54]. It is been argued that the use of oversampling techniques could lead to overly optimistic prediction results [55]. Therefore, in order to deal with an unbalanced data set, we decided to use weight class balancing. Table 5 provides detailed information of the parameters optimization process for the RF using Scikit Learn's GridSearchCV function [49].

Table 5 Parameter optimization for RF

| Parameter | Tested values |
| --- | --- |
| Number of trees | From 50 to 1000 (step length =50) |
| Number of variables for each split | 2,6,8,10 |
| Minimum number of samples required to be at a leaf node | $1\times10^{-2}$, $1\times10^{-3}$, $1\times10^{-4}$, $1\times10^{-5}$, $1\times10^{-6}$ |
| Minimum impurity required to split a node | $1\times10^{-2}$, $1\times10^{-3}$, $1\times10^{-4}$, $1\times10^{-5}$, $1\times10^{-6}$ |

Lastly, NNs are widely recognized for their capability to model complex statistical interactions between variables [56,57]. An NN is a system of interconnected neurons, organized by independent layers, inspired by biological nervous functioning [58]. Each neuron accepts a number of weighted inputs and processes them to produce an output [59]. The weights of the network connections measure the potential amount of the knowledge of the network [60]. Therefore, a training phase is needed in which the NN adapts the weights through minimization of the error between actual and estimated outputs [58].

NNs have been shown to be highly accurate for classification problems. However, the major drawback is that they are considered black-box models [52]. Their nested non-linear structure makes it difficult to understand what information in the input data makes them arrive at their decisions [61]. Recently, Layer-wise Relevance Propagation (LRP) has been proposed as a general solution to the problem of understanding classification decisions [62]. The algorithm relies on a conservation principle to propagate the prediction back throughout the network, ensuring the network output is fully redistributed through the layers of the NN back to the input variables [63]. The main idea is to understand which input variables contribute to a positive or negative classification result [62]. Recent applications of LRP include sentiment analysis [64] and image classification [65]. However, to the best of our knowledge, this explainable approach has not been used in the prediction of no-show probabilities. This is particularly relevant in our case, since lack of explainability in decision support systems can lead to both practical and ethical issues [66].

According to Guo and Berkhahm [67], the continuous nature of NNs limits their applicability to categorical variables. Although one-hot encoding is a popular approach to overcome such limitations, it can require an unrealistic amount of computational resource, increase variance and ignore informative relationships between variables. To deal with this problem, Guo and Berkhahm [67] apply the logic used in natural language processing and design an entity embedding method for categorical variables. The idea is to map discrete values to a multi-dimensional space where values with a similar function output are close to each other. Since the new representation increases the continuity of the data, it speeds up the training process and exploits intrinsic properties of categorical variables. For this work, both one-hot encoding and categorical embeddings are tested. Therefore, we implement an NN with one hidden layer and use heat maps produced by LRP to identify features supporting the classifier's decision for or against a specific class [64]. Table 6 provides basic information regarding parameters optimization using Scikit Learn's GridSearchCV function [49]. All other parameters have been left at their default values.

Table 6 Parameter optimization for NN

| Parameter | Tested values |
| --- | --- |
| Number of iterations | From 100 to 1600 (step length =100) |
| Number of neurones | 10 values from N/2 to 2N, where N is the number of variables |

3.4 Demonstration and Evaluation

Models performance was assessed using the AUROC score. From the available database, we randomly generated training (70%) and test (30%) sets. In the demonstration phase, a 10-fold Cross Validation process repeated 10 times (10-by-10 CV) was carried out using the training set. In the evaluation phase, we used the test set to assess the quality of the results and discuss the practical implications of increased accuracy when implementing interventions to reduce no-shows. A public version of the experimentation code and a randomly generated database are available [68]. Different patients may be selected for targeting according to which classification algorithm is used. As discussed in Section 3.2, we use two measures to assess the quality of a given classification, coverage and risk. Therefore, with this evaluation approach, we aim at quantifying the potential impact of using our DSS.

**4. Data collection and Initial Analyses**

We analyse data from the South West cluster of the city. Forty-nine medical facilities offering primary care are located within this cluster. Four services: Oral Health (OH), Growth and Development (G&D), Young Adult Program (YAP) and Senior Program (SP), are studied as these cover 75% of scheduled appointments during 2017 and 2018. In many scoring models, segmentation of discrete variables results in more stable and parsimonious models [69], so we have used decision trees to classify these categorical variables coarsely (for age and lead time) and one-hot encoding to represent them in the models. Table 7 presents a list of patient and appointment-level variables, their descriptions and the Cramer's V correlation coefficient with the outcome (show or no-show).

Table 7 List of variables

| Category | Variable | Description | Correlation Coefficient | | | |
|---|---|---|---|---|---|---|
| | | | OH | G&D | YAP | SP |
| Patient | Gender | Gender of the patient (Male, Female) | 0.000 | 0.000 | 0.000 | 0.000 |
| | Age | Age of the patient at the moment of the appointment (years) | 0.002 | 0.001 | 0.002 | 0.000 |
| | Zone | Area of the city where the patient lives | 0.008 | 0.017 | 0.008 | 0.008 |
| Appointment | Lead time | Elapsed time between the date of the home visit and the appointment date (days) | 0.015 | 0.018 | 0.008 | 0.005 |
| | Month | Month in which the appointment was scheduled | 0.009 | 0.015 | 0.012 | 0.009 |
| | Day | Day of the week in which the appointment was scheduled | 0.006 | 0.005 | 0.006 | 0.001 |
| | Facility | Assigned healthcare facility | 0.009 | 0.037 | 0.005 | 0.008 |

Our analysis also uses publicly available information relating to two of the above variables. First, using data from the National Administrative Department of Statistics (*Departamento Administrativo Nacional de Estadística*, DANE), we classify the zone in which each patient lives as low-income if 50% or more of its population belongs to the lowest two income strata: otherwise, it is classified as medium-income. We also use data provided by the National Police Office [70] on reported criminal events affecting individual citizens since 2015 to determine the sociodemographic context of each healthcare facility.

Table 8 summarizes basic information on 53,311 scheduled appointments during these two years, including the outcome (show or no-show). Oral Health has the greatest number of appointments (22,613, 42%) and Growth and Development the least, at 14%. No-show rates range from 21% to 39% for each service. At an aggregate level, there is no difference in gender between the no-show rates, but in Oral Health and Young Adult Program more females than males keep their appointments. With respect to age, the highest no-show rates are between 20 and 40 years and the lowest are among children under 10 years and adults over 50. It is also possible to see that smaller lead times yield lower no-show rates. The only exception to this behaviour is in Senior Program where no-show rates are slightly lower for appointments assigned more than 60 days in advance. Finally, Figure 1 shows the attendance levels, for each service, presented by day of the week and month of the year. While, on average, 92% of the patients keep their appointments on Sunday, this indicator decreases to 69% on Fridays. Attendance levels range from 58% to 82%, for each month, and its behaviour changes across the four services.

Table 8 Descriptive statistics

| Category | OH | | | G&D | | | YAP | | | SP | | | **Total** | | |
|---|---|---|---|---|---|---|---|---|---|---|---|---|---|---|---|
| Gender | Show | No-show | | Show | No-show | | Show | No-show | | Show | No-show | | **Show** | **No-show** | |
| Women | 8,457 | 3,475 | **29%** | 2,629 | 965 | **27%** | 4,078 | 1,946 | **32%** | 3,294 | 1,067 | **24%** | **18,458** | **7,453** | **29%** |
| Men | 7,482 | 3,199 | **30%** | 2,650 | 999 | **27%** | 4,035 | 2,065 | **34%** | 5,365 | 1,605 | **23%** | **19,532** | **7,868** | **29%** |
| Age (years) | | | | | | | | | | | | | | | |
| 0-10 | 1,659 | 638 | **28%** | 5,197 | 1,918 | **27%** | 0 | 0 | -- | 0 | 0 | -- | **6,856** | **2,556** | **27%** |
| 10-20 | 2,860 | 1,186 | **29%** | 82 | 46 | **36%** | 5,856 | 2,530 | **30%** | 0 | 0 | -- | **8,798** | **3,762** | **30%** |
| 20-30 | 1,467 | 789 | **35%** | 0 | 0 | -- | 2,092 | 1,178 | **36%** | 0 | 0 | -- | **3,559** | **1,967** | **36%** |
| 30-40 | 2,316 | 1,058 | **31%** | 0 | 0 | -- | 165 | 91 | **36%** | 0 | 0 | -- | **2,481** | **1,149** | **32%** |
| 40-50 | 2,964 | 1,209 | **29%** | 0 | 0 | -- | 0 | 0 | -- | 1,294 | 425 | **25%** | **4,258** | **1,634** | **28%** |
| 50-60 | 2,093 | 791 | **27%** | 0 | 0 | -- | 0 | 0 | -- | 3,232 | 1,014 | **24%** | **5,325** | **1,805** | **25%** |
| Over 60 | 2,580 | 1,003 | **28%** | 0 | 0 | -- | 0 | 0 | -- | 4,133 | 1,233 | **23%** | **6,713** | **2,236** | **25%** |
| Lead time (days) | | | | | | | | | | | | | | | |
| 0-15 | 7,689 | 2,259 | **23%** | 3,329 | 929 | **22%** | 4,969 | 2,029 | **29%** | 4,070 | 1,068 | **21%** | **20,057** | **6,285** | **24%** |
| 15-30 | 2,079 | 1,061 | **34%** | 754 | 453 | **38%** | 902 | 562 | **38%** | 1,013 | 434 | **30%** | **4,748** | **2,510** | **35%** |
| 30-60 | 1,503 | 785 | **34%** | 373 | 166 | **31%** | 488 | 299 | **38%** | 901 | 309 | **26%** | **3,265** | **1,559** | **32%** |
| Over 60 | 4,668 | 2,569 | **35%** | 823 | 416 | **34%** | 1,754 | 1,121 | **39%** | 2,675 | 861 | **24%** | **9,920** | **4,967** | **33%** |

OH: Oral Health, G&D: Growth and Development, YAP: Young Adult Program, SP: Senior Program

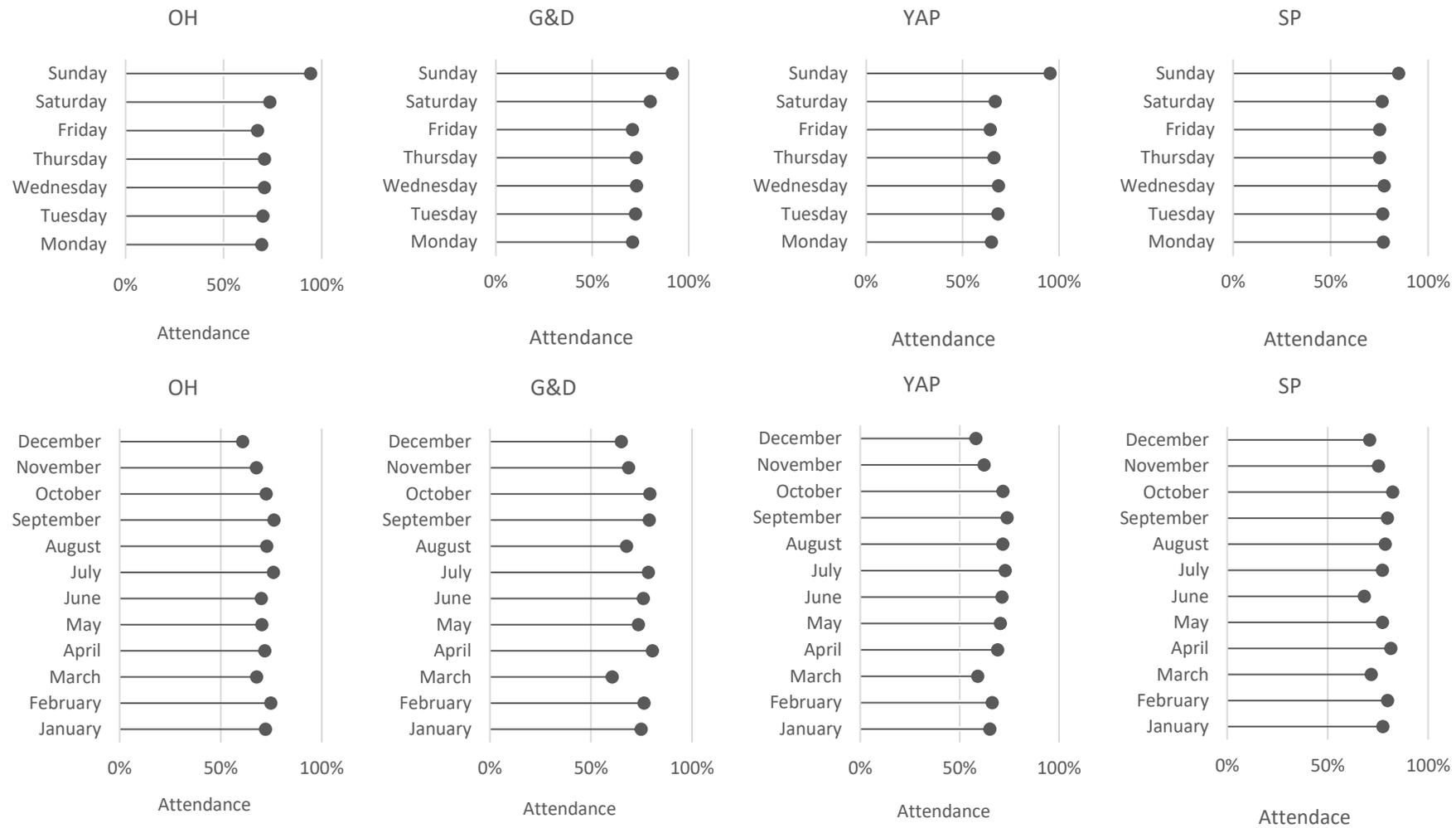

Figure 1 Attendance levels for each service

# 5. Results and discussion

We present the results organized in three sections. Firstly, we quantify the impact of each variable in the LASSO regression model on the no-show probability, and analyse the average coefficients obtained by a 10-by-10 cross validation process. Next, we quantify the added value of using RF and NN and compare the four modelling alternatives using the AUROC score. Finally, we analyse the impact of using these results as a decision support system and quantify changes in coverage and risk of an intervention when accuracy of prediction models increases.

5.1 LASSO regression model: variables affecting no-show probabilities

Females are more likely to keep their appointments except for SP (odds ratio OH: 1.03, G&D: 1.01, YAP: 1.08 and SP: 0.95). On the one hand, this result is highly context-dependent. Whereas some studies have reached the same conclusion [71,72], others have reported that males have lower no-show rates [19,73,74], or concluded that gender does not have impact in no-show probabilities [75,76]. On the other hand, in developing countries it has been argued that, among socio-economically disadvantaged females, high no-show rates might be related to a lack of support from social networks and their responsibilities as caregivers [77–79]. This might explain the result for SP. In Bogotá it is common to find that women older than 60 years take care of their grandchildren while the parents work.

No-show probabilities change with the age of the patient. Figure 2 shows odds ratios (OR) for each age range in the four services. Since four models were run, one for each service, the reference values (OR = 1) must be independently interpreted. Firstly, in OH, patients between 22 and 33 years have the highest no-show probability (OR = 0.72) and it is not possible to identify any age range in which patients have particularly low no-show rates. Secondly, in G&D (between 0 and 13 years) and YAP (between 14 and 44 years), older patients are more likely to miss their appointments. This result is consistent with previously reported findings in primary care and paediatrics settings [20,73,76]. Finally, age seems to have less impact among SP patients.

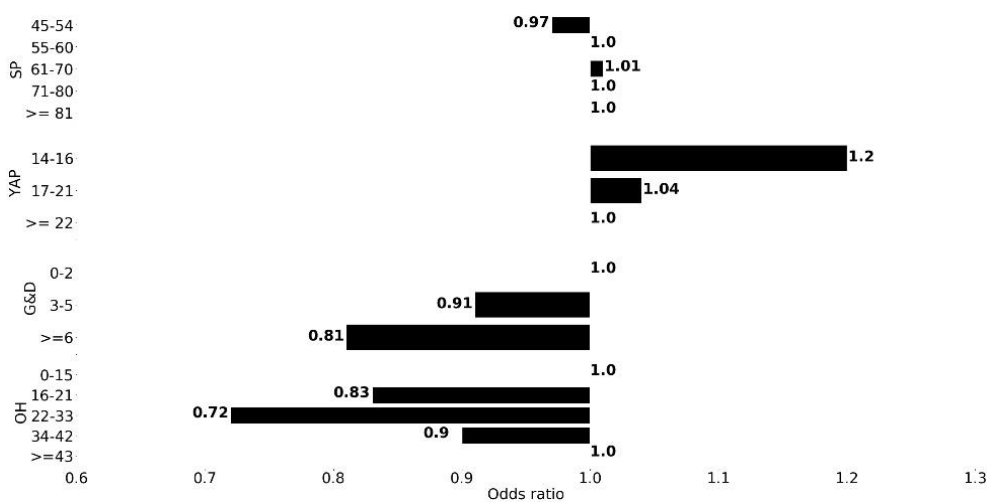

Figure 2 Odds ratio for each range of age

There is a relationship between the estimated income of the zone in which the patient lives and the no-show probability. In OH, 67% of zones with OR lower than one (i.e. higher no-show probabilities), are low-income zones. Additionally, 63% of zones with lower no-show probabilities have middle-income levels. This result might be associated with the patient's perception of oral health needs. Wallace et al. [80] found that, among low-income populations, dental care is perceived as desirable but more as a luxury than a necessity. The opposite scenario was found in SP: 75% of the zones in which OR is less than one and 25% of the zones with OR greater than one, have medium-income levels. Finally, for G&D and YAP, low-income zones represent the majority of both low and high no-show probability groups.

As expected, longer lead times increase no-show probabilities. Similar results have been reported in primary care settings [19,20], and paediatric clinics [15]. As can be seen in Figure 3 for G&D and YAP, patients who are scheduled for appointments with lead times between 8 and 10 days are more likely to attend. Given the age of patients in these services, a companion (parent or carer) is often required to attend the appointment. Consequently, non-attendance may be due to challenges in coordinating these logistical aspects. On the other hand, the best attendance rates in OH occur when the lead-time ranges from 0 and 10 days, and the probability of no-shows reaches its maximum value after 15 days. Finally, in older patients, the probability of attendance changes less with respect to lead time. These results demonstrate the non-linear nature of no-shows and support the use of analytical techniques for scheduling appointments.

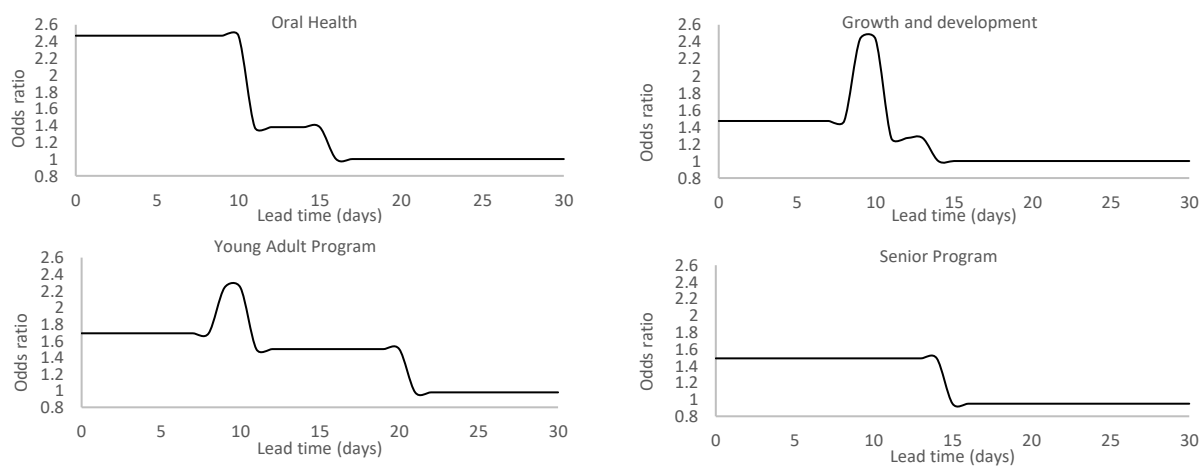

Figure 3 Odds ratio for each service when lead time is varied

We also find the date of the appointment affects the no-show probability. Previous studies have reported seasonal behaviours through the year [71,73] or variations depending on the day of the week [28,81,82]. Table 9 shows the OR variations, for the four services, in each month of the year and each day of the week. As can be seen, in YAP the probability of no-show changes slightly in January, March and October. However, for the other months of the year and for all the days of the week, it remains constant. On the other hand, the service with the most changes throughout the year is SP with odds ratio varying between 0.61 and 1.22. This result is interesting because no-show probabilities show low sensitivity to factors such as age, area and lead-time in this service. The months of March, November and December seem to have higher levels of risk of no show. January, February and October have better behaviour. Regarding the days of the week, the best attendance levels are seen during the weekend.

Finally, we observe that there is a relationship between neighbourhood crime statistics and no-show probabilities. The four healthcare facilities with OR lower than one, across the four services, are in neighbourhoods with the highest number of incidents. Similarly, out of 49 facilities used in the program only two have OR greater than one, across all four services. These facilities are in neighbourhoods with the lowest incidence of crime in their respective districts.

Table 9 Average odds ratio for the appointment date

| Month of the year | G&D  | YAP  | SP   | OH   |
|---|---|---|---|---|
| January   | 1.02 | 0.96 | 1.01 | 1.11 |
| February  | 1.10 | 1.00 | 1.22 | 1.30 |
| March     | 0.72 | 0.85 | 0.87 | 1.00 |
| April     | 1.00 | 1.00 | 1.13 | 1.00 |
| May       | 1.00 | 1.00 | 0.99 | 0.80 |
| June      | 0.9  | 1.00 | 0.61 | 0.75 |
| July      | 1.02 | 1.00 | 0.99 | 1.04 |
| August    | 0.74 | 1.00 | 1.00 | 0.90 |
| September | 1.00 | 1.00 | 1.00 | 1.00 |
| October   | 1.04 | 1.01 | 1.21 | 1.00 |
| November  | 0.71 | 1.00 | 0.94 | 0.95 |
| December  | 0.51 | 1.00 | 0.73 | 0.69 |
| Day of the week | G&D | YAP | SP | OH |
| Sunday    | 1.06 | 1.00 | 1.37 | 3.50 |
| Monday    | 0.96 | 1.00 | 1.00 | 0.93 |
| Tuesday   | 1.00 | 1.00 | 1.00 | 0.96 |
| Wednesday | 1.00 | 1.00 | 1.00 | 1.00 |
| Thursday  | 1.00 | 1.00 | 0.96 | 1.00 |
| Friday    | 0.96 | 1.00 | 0.98 | 0.84 |
| Saturday  | 1.35 | 1.00 | 1.00 | 1.03 |

Figure 4 presents the 15 variables with the greatest impact on no-show probabilities in each service. Firstly, attendance levels are higher on Saturdays (G&D) and Sundays (OH, YAP and SP). This result can be used to inform tactical decisions regarding the number of appointments that should be made available throughout the planning horizon. Secondly, ensuring reasonable lead times can affect utilization levels. For the four services, it is possible to identify a maximum lead time that can be set as an objective in the scheduling process. Lastly, some months and facilities have relatively high no-show rates. This information can be used in the design of overbooking policies, since better estimates of no-show probabilities would help to reduce the undesirable side effects of this practice.

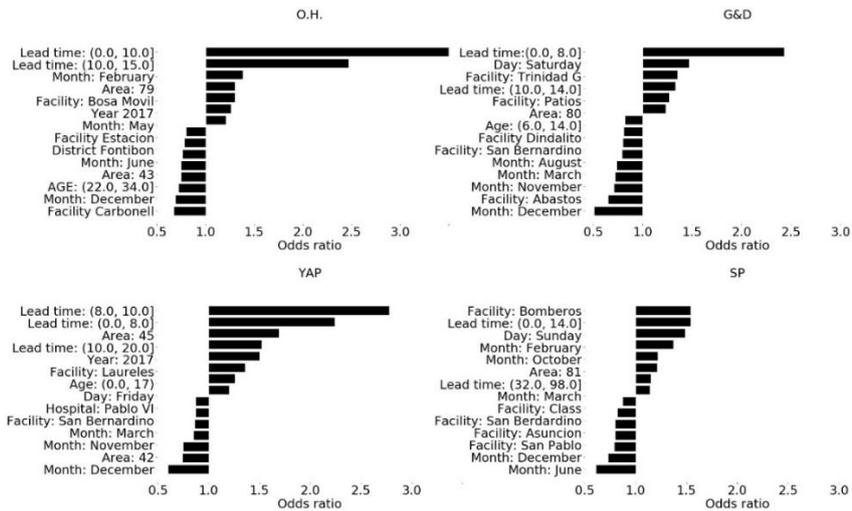

Figure 4 LASSO results: most relevant variables for each service

5.2 The added value of using other modelling approaches

Figure 5 presents the AUROC performance of the four models. Each point in the graph represents the average and standard deviation of the AUROC in a repetition of 10-by-10 cross-validation. For all four services, the NN model has better average performance. The difference between RF and LR may suggest that the non-linear component of the relationship between the variables is not very strong, but still significant. Additionally, the average AUROC of both NN models indicates that variable interaction can be successfully modelled without using categorical embedding. Despite having low standard deviations, the amount of available data might not be sufficient to generate robust embeddings, and the disadvantage of increased variance is outweighed by better bias estimation with the increased number of dummy variables. Therefore, in our further analysis we consider only the NN with one-hot encoding.

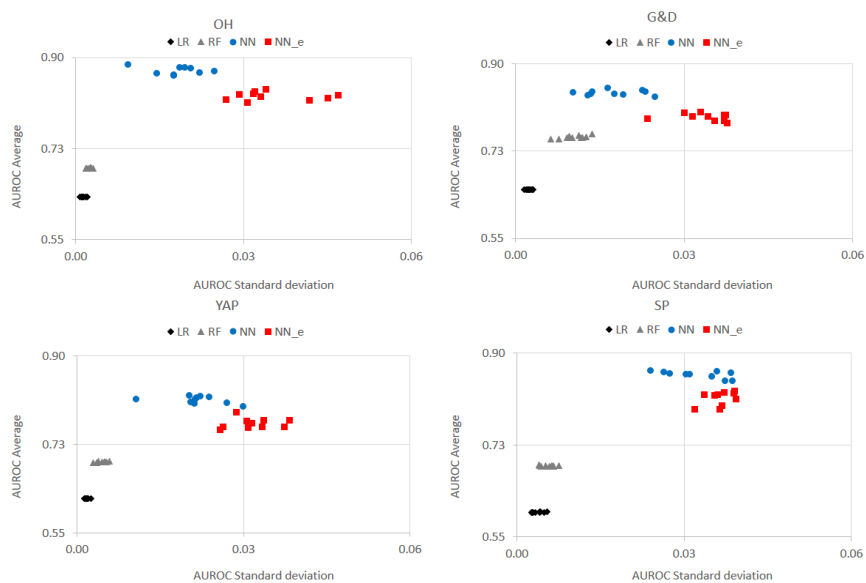

Figure 5 Model performance

5.3 The decision support system

In this section we analyse the implications of using our DSS to select patients and implement targeted interventions. As discussed in Section 3, SDS have specified that the developed DSS should allow program managers to divide patients into three groups. Group A will contain the 30% of patients estimated to have lowest no-show risk. Group B, the 40% of patients with intermediate risk and Group C will contain the remaining 30% of patients with the highest no-show probabilities. From the machine learning perspective, this means that two cut-off points are required. This process is called cut-off point tuning and is based on ROC performance measures [83].

Table 10 presents the coverage and risk for each service and for three models. As can be seen, in G&D, an intervention for 30% of patients could cover 80% of the no-shows, if the prediction of the NN is used. This percentage would decrease to 64% or 41% using RF or LR. This result is consistent across the four services. The average improvement in coverage using NN is 14% with respect to RF and 62% with respect to LR. Lastly, the risk of not implementing any action among patients with low no-show risk can be quantified. Using the NN classification, these patients represent between 2% and 3% of no-shows. On the other hand, using the RF prediction the highest risk is in OH and YAP where 9% of the no-shows are classified in group A. Finally, using logistic regression, the risk varies between 17% and 21%.

Table 10 Risk and Coverage for a potential intervention

| Service | NN Risk | NN Coverage | RF Risk | RF Coverage | LR Risk | LR Coverage |
|---|---|---|---|---|---|---|
| OH | 2% | 70% | 9% | 55% | 17% | 39% |
| G&D | 2% | 80% | 3% | 64% | 17% | 41% |
| YAP | 3% | 67% | 9% | 55% | 20% | 41% |
| SP | 2% | 75% | 6% | 61% | 21% | 40% |

Despite the high accuracy of the NN prediction (i.e. risk of 2.25% and average coverage around 72%), using NN results might be challenging. It has been argued that decision makers need to understand the reasons underpinning a prediction in order to trust the results [84,85]. Consequently, in an attempt to explain the results of the NN, we implement LRP [62]. As a result, the importance of each of the variables in the classification of a patient in each category is obtained. According to Yang, Tresp, Wuderle and Fasching [65], one of the advantages of this technique, compared with sensitivity analysis, is the interpretability of the signs (and absolute values) of the weights. For example, when the weight is large and positive, the variable strongly supports the classification chosen by the NN, whereas if the weight is small and negative, the variable weakly suggests the opposite classification.

Figure 6 illustrates the results obtained for ten G&D patients. A column represents a patient, and the column heading the no-show probability. The blue cells represent positive coefficients and the red cells negative ones, shaded by the magnitude of the weight. Only those variables with at least one non-zero coefficient are shown for this group of patients. Our NN predicts that the first patient will not show up for his appointment (95% of no-show probability). The main reason for this conclusion is the month of the appointment, but lead time and age also support this classification. On the other hand, the day of the appointment, and the zone in which the patient lives support the opposite classification, i.e. that the patient would in fact attend. Moreover, the fact that the same variables have both positive and negative coefficients (regardless of the no-show probabilities) implies that the network is learning about the context. It means that the NN learns that it is not enough to know the gender of a patient to decide in which category they should be classified. For example, when running a regression model including the interaction of gender with age and day of the week, it is possible to observe slightly different results from those reported by the model without interaction. For G&D we found that females are more likely to miss appointments on Saturdays and Wednesdays (O.R. = 0.75 and 0.94).

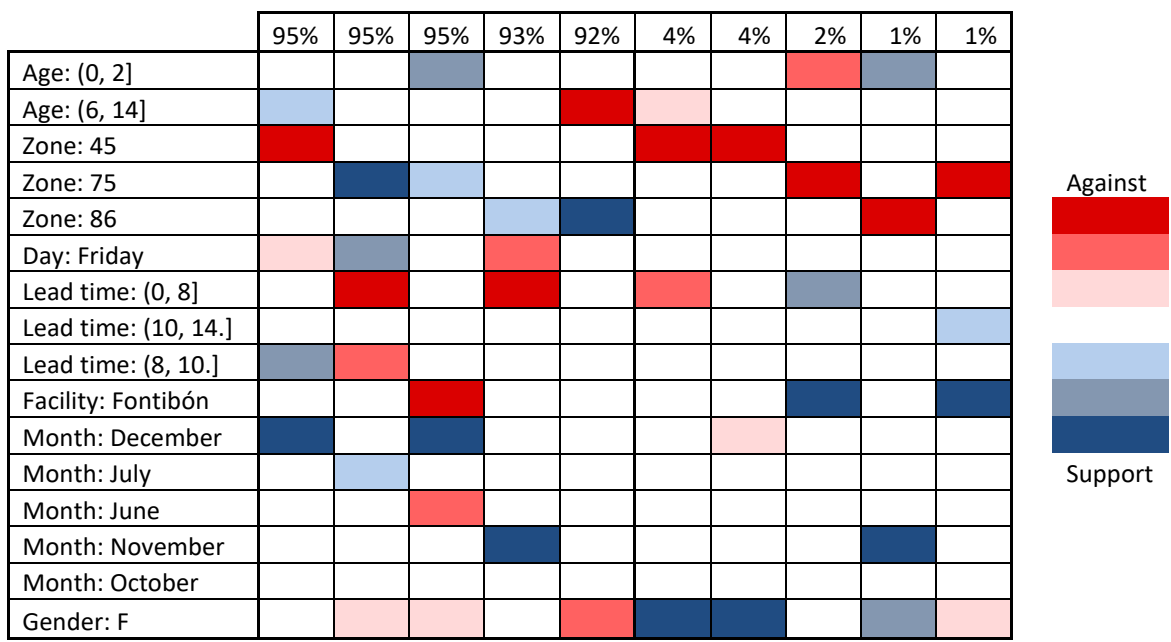

Figure 6 Heatmap for G&D

This interpretation procedure can be used for any data-driven machine learning model, giving interesting insights which have real applications for decision support systems. The focus of the implementation of these processes must be on actionable combinations of parameters: if a relation between several variables is found (such as gender and day of the week), then an action can be performed for those specific groups. The use of such interpretability tools allows effects like these to be observed and revealed, which would otherwise be neglected.

**6. Concluding remarks**

This work has been developed with the active involvement of SDS. We conducted several workshops with program managers, operational analysts at the healthcare clusters, and community workers. This approach enabled us to develop a more comprehensive view of the program, identify a champion for the project within SDS, generate a shared vision of the main challenges and maintain stakeholders' engagement. Therefore, the discussion shifted from a resource allocation perspective to a better understanding of the underlying problem and the identification of restrictions that could prevent the implementation of changes in the operation. By the end of the workshops, they concluded that as the program covers a highly heterogenic population, more personalized strategies were required to reduce no-show behaviour. In this context, patient classification was identified as a key element to ensure the economic feasibility of any intervention strategy. Following this process we were able to reach an agreement on an appropriate design principle which captures the knowledge of our stakeholders. Consequently, our DSS aims at leveraging routinely collected data to inform such classification. During the feedback sessions, program managers stressed that they value that the results were easy to understand and have the potential to improve service quality.

In light of the promising results, at the time of writing SDS is starting a pilot intervention to modify patient behaviour. Using the DSS to predict no-show probabilities will improve the effectiveness of the intervention since the NN classifies around 80% of potential no-show cases to Group C, the 30% of patients designated to receive the most intensive, personalized, intervention. Therefore, as stated in our design principle, the design of this DSS can be seen as a necessary first step to reduce no-show behaviour in primary care and the future development of design knowledge [86]. After an evaluation of the pilot is carried out, a web-based tool will be developed to enable models to be used in program operation.

Our objective is to use routinely collected data to predict no-show probabilities. However, in order to inform the discussion, we also use other sources of public information. Two findings are particularly relevant in developing country contexts. First, in Oral Health, 67% of zones with OR less than one (i.e. higher no-show probabilities), are low-income zones, and 63% of zones with lower no-show probabilities have middle-income levels. Second, the four healthcare facilities (out of 49) with OR less than one across all four services are in neighbourhoods with the highest number of reported crime incidents, whereas the two facilities with OR greater than one across all four services are in neighbourhoods with the lowest crime incidence in their respective districts. These exploratory results indicate that including socioeconomic data could potentially increase understanding of no-show behaviour. Further research is required to evaluate the cost/benefit of collecting such data and including these variables in the SDS information system.

Since this paper presents retrospective analysis of historical data, it is not possible to draw conclusions regarding the reasons for no-show behaviour. Therefore, the next step in this research is a mixed-method study to understand and model patients' decision-making processes. Semi-structured interviews are being conducted among no-show patients in order to learn from their experiences and identify access barriers. Moreover, we are analysing survey data to quantify the relationship between the no-show probability and the constructs of a health psychology model.

Modelling no-show behaviour in primary care settings is an active research area. The most common approach in the literature is multiple regression, but due to its limitations in accuracy, using the results to improve management practice can be problematic. Machine learning methods are gaining in popularity but have the drawback of being a 'black box' that managers may not trust. Our research shows the benefits of a two-pronged approach to overcome this. Firstly, using LRP to produce a heat map visualisation that makes the model results immediately understandable, and secondly, involving stakeholders in every stage of the design process. Additionally, despite the highest no-show rates worldwide being in developing countries, the problem has been mainly studied in North America and the United Kingdom [7]. The research presented here contributes to the literature by assessing the effectiveness of machine learning approaches using routine data to predict no-show behaviour among low-income patients in a developing country context.

Throughout this research, we have designed a process to construct a DSS for improving healthcare access powered with machine learning models. In general, the first stages of the process are replicable across all healthcare systems that wish to develop such a DSS. Steps such as the facilitated formulation of a design principle, data collection strategies, data processing steps, and the selection of models to use are relevant issues for every designer facing a similar challenge. Additionally, by using the schema proposed by Gregor et al. [45], our design principle also meets the five criteria for reusability: accessibility, importance, novelty, and effectiveness [87]. Therefore, we expect that other DSS designers can build from this experience when defining intervention groups. We believe that our results could help not only other designers in healthcare settings but also those dealing with limited resources in other contexts such as educational or social programs, where prioritization plays a key role in ensuring feasibility.

**Funding**

The first author's research is partially funded by a PhD scholarship from the healthcare research stream of the program *Colombia Científica – Pasaporte a la Ciencia*, granted by the Colombian Institute for Educational Technical Studies Abroad (Instituto Colombiano de Crédito Educativo y Estudios Técnicos en el Exterior, ICETEX). The third author acknowledges this research was undertaken, in part, thanks to funding from the Canada Research Chairs program.